\documentclass[final,5p,times,twocolumn,authoryear]{elsarticle}

\usepackage{enumitem}

\usepackage{caption}
\PassOptionsToPackage{hyphens}{url}
\usepackage[colorlinks = true, linkcolor = blue, urlcolor = blue,           citecolor = blue, anchorcolor = blue]{hyperref}

\usepackage{moresize}

\begin{document}

\begin{frontmatter}



\title{Comprehensive analysis of synthetic learning applied to neonatal brain MRI segmentation}


\author[first]{R. Valabregue}
\author[first]{F. Girka}
\author[sec]{A. Pron}
\author[ter]{F. Rousseau}
\author[sec]{G. Auzias}
\address[first]{CENIR, Institut du Cerveau (ICM) - Paris Brain Institute, Inserm U 1127, CNRS UMR 7225, Sorbonne Université, Paris, France}
\address[sec]{Aix-Marseille Université, CNRS, Institut de Neurosciences de la Timone, UMR 7289, Marseille, France}
\address[ter]{IMT Atlantique, LaTIM INSERM U1101, Brest, France}

\begin{abstract}
Brain segmentation from neonatal MRI images is a very challenging task due to large changes in the shape of cerebral structures and variations in signal intensities reflecting the gestational process. In this context, there is a clear need for segmentation techniques that are robust to variations in image contrast and to the spatial configuration of anatomical structures. In this work, we evaluate the potential of synthetic learning, a contrast-independent model trained using synthetic images generated from the ground truth labels of very few subjects.\\
We base our experiments on the dataset released by the developmental Human Connectome Project, for which high-quality T1- and T2-weighted images are available for more than 700 babies aged between 26 and 45 weeks post-conception. First, we confirm the impressive performance of a standard Unet trained on a few T2-weighted volumes, but also confirm that such models learn intensity-related features specific to the training domain. We then evaluate the synthetic learning approach and confirm its robustness to variations in image contrast by reporting the capacity of such a model to segment both T1- and T2-weighted images from the same individuals. However, we observe a clear influence of the age of the baby on the predictions. We improve the performance of this model by enriching the synthetic training set with realistic motion artifacts and over-segmentation of the white matter. Based on extensive visual assessment, we argue that the better performance of the model trained on real T2w data may be due to systematic errors in the ground truth. We propose an original experiment combining two definitions of the ground truth allowing us to show that learning from real data will reproduce any systematic bias from the training set, while synthetic models can avoid this limitation. Overall, our experiments confirm that synthetic learning is an effective solution for segmenting neonatal brain MRI. Our adapted synthetic learning approach combines key features that will be instrumental for large multi-site studies and clinical applications.
\end{abstract}




\end{frontmatter}




\section{Introduction}
\label{introduction}
\subsection{Context}
Automated segmentation of perinatal brain MRI remains a challenging task due to massive changes in the global shape of the brain and large variations in image intensity reflecting the rapid tissue maturation that occurs around birth \cite{kostovic_neural_2019}. Several segmentation methods specifically designed for increased robustness to these factors such as multi-atlas label fusion techniques have been proposed \cite{makropoulos_automatic_2014}, \cite{gholipour_multi-atlas_2012}, \cite{benkarim_toward_2017, makropoulos_review_2018,li_computational_2019}. Those methods have been applied to large open datasets like the developing Human Connectome Project  (dHCP)  \cite{makropoulos_developing_2018} enabling a better characterization of early brain development \cite{edwards_developing_2022,dimitrova_phenotyping_2021}.\\
More recently, supervised deep-learning techniques have been introduced as the next generation of segmentation techniques in medical images, showing higher performances and lower computing time than previous approaches. In particular, the UNet architecture \cite{ronneberger_u-net_2015} outperformed previous approaches in many different challenges \cite{isensee_nnu-net_2021}. Nevertheless, a well-known limitation of supervised learning methods is their strong reduction in performances when applied to unseen data \cite{karani_lifelong_2018}. This “domain-gap” problem has been identified as a major bottleneck in the field \cite{pan_survey_2010, zhou_domain_2022} and an extensive body of literature investigated potential solutions and reported various gains in robustness, depending on the context. While our focus is not to review this large literature, we summarise the main approaches in order to better situate our strategy in the context of perinatal brain MRI segmentation. \\
A common approach consists in augmenting the training set with synthetic perturbations that explicitly control for the deviation from the initial training dataset \cite{ilse_selecting_2021}. An obvious advantage is that it avoids the costly solution of getting more training data. In the context of brain development, two aspects of data augmentation corresponding to the two key challenges pointed above can be distinguished: 1) \textbf{spatial} augmentation to account for variations in the spatial arrangement of the different tissues and in the shape of specific anatomical structures (e.g., increase in cortical folding with age);  2) \textbf{style} (or appearance) augmentation to account for changes in tissue contrast, which can be induced either by variations in the acquisition settings or scanner or by variations related to brain maturation. \\
The design of those synthetic augmentations can either rely on a \textbf{physics-based} (i.e., with a direct analytic model) or a \textbf{learning-based} generative model. Examples of physics-based augmentation strategies combine random affine or nonlinear deformations for spatial augmentation and random gamma perturbations for intensity augmentation \cite{zhang_generalizing_2020, perez-garcia_torchio_2021}. The efficiency of this type of approach has been demonstrated in an intra-modality context \cite{zhao_data_2019,zhang_generalizing_2020, isensee_nnu-net_2021}, but it is less efficient for cross-modality \cite{karani_lifelong_2018}. The learning-based augmentation techniques are often referred to as “domain adaptation”. The recent works in this field have focused on the design of unsupervised learning approaches aiming at generating realistic synthetic training sets without requiring manually labeled data in the target external domain. Such techniques either learn a latent space that is common to the original domain where ground truth labels are available and to the target external domain \cite{kamnitsas_unsupervised_2017, csurka_domain-adversarial_2017,tomar_self-supervised_2022} or learn a direct image-to-image translation \cite{zhang_translating_2018}. These two approaches are combined in \cite{chen_synergistic_2019}. The use of adversarial generative models for domain adaptation has also been considered in \cite{chartsias_multimodal_2018}. \\
Another key feature of synthetic augmentation techniques is the capacity to train these models using very few manually labeled training data. Indeed, one-shot learning studies propose to reduce the training data to only one template image with corresponding ground truth labels \cite{tomar_self-supervised_2022, zhao_data_2019}. All those methods alleviate the need for time-consuming and expertise-demanding ground truth segmentation in the target domain since the training of the domain transfer model requires a pool of unlabeled data representative of the target domain. The major drawback of the learning-based approaches is the need to train a new model for any new domain. Of note, the very recent work  \cite{ouyang_causality-inspired_2022} proposes to leverage this limitation by generating a wide range of contrasts from a single domain dataset using augmentation techniques inspired by the different acquisition processes.\\
Recently, Billot et al. introduced a method called SynthSeg \cite{billot_synthseg_2023, billot_learning_2020} that does not rely on any real MRI data during the training process. We refer to this type of model as "\textbf{synthesis-based}". The key is to avoid the potential bias toward the domain of the training set by introducing a framework allowing to train the models without any real imaging data. A fully synthetic training dataset is generated from a set of real labels maps using physics-based generative models of the correspondence between label maps geometry and underlying intensity distributions. Under the assumption of homogeneous tissues, the image signal is sampled from a Gaussian distribution with different mean and variance for each tissue (MR data is indeed a mixture of Gaussian intensities). The generated signal is then enriched with additional commonly used random transformations (bias field, gaussian noise, and spatial deformation).\\
The approach proposed by Billot et al. is based on the “domain randomization” concept \cite{tobin_domain_2017, tremblay_training_2018}, which postulates that the variations across real data from different domains need to be encompassed within the distribution of the generated synthetic data. Therefore, the setting of the parameters in the generative process is key, and the design of the transforms has to generate large enough variations, without strong constraint on their biological relevance. More specifically in the context of the present work, robustness to variations in image intensity distributions can be favored by randomly sampling the mean and variance of the generated signal; while the robustness to variations in brain size and cortical folding magnitude can be induced by tuning the random deformations. In \cite{billot_synthseg_2023, billot_robust_2022}, the authors validated their approach on highly heterogeneous data acquired on adults using various settings from clinical practice, demonstrating impressive robustness to challenging variations in image contrast and resolution. They reported higher segmentation accuracy and robustness compared to other methods of domain adaptation. In addition, the authors of \cite{billot_synthseg_2023} investigated the influence of the size of the training set on the performances of SynthSeg and reported that only a few training examples are sufficient to converge towards its maximum accuracy on a population of adults. In the present work, we assess whether these conceptually appealing features and impressive results on the adult population extend to the context of neonatal brain MRI segmentation.

\subsection{Contributions}
Our work focuses on a comprehensive analysis of synthetic learning approaches for segmenting neonatal brain MRI data. To this end, we first reimplemented the SynthSeg model \cite{billot_synthseg_2023} within the Pytorch framework relying on the torchio transformations \cite{perez-garcia_torchio_2021}. The code is made available \footnote{\url{https://github.com/romainVala/torchQC/tree/master/segmentation}}. In contrast to \cite{billot_synthseg_2023} who demonstrated the robustness of SynthSeg to variations in image resolution and contrast on a very large clinical dataset, we focus here on the potential advantages when applied to perinatal brain MRI, in comparison to a classical UNet trained on real T2w data using a few shot learning strategy. Since the SynthSeg model did not perform as well as expected on neonatal brain MRI data, we propose two solutions to address its limitations, yielding better performances: adding simulated motion augmentation or subdividing the WM tissue into several sub-compartments.\\
Using our improved synthesis-based model, we then confirm the robustness of the predictions to variations in the contrast of the images, with very consistent predictions from either T1w or T2w images from the same subjects. We also demonstrate another key advantage of the synthetic learning approach: the synthesis-based models learn an unbiased correspondence between the geometry of the labels and image intensities. To quantitatively support this feature, we build a second type of ground truth from the same dataset. We report a much lower influence of the definition of the ground truth on the predictions from the synthesis-based models, compared to a model learned on real MRI data, which reproduces any systematic bias from the ground truth.\\
The quantitative evaluations are complemented with a careful visual assessment of the predictions and ground truth. This allows us to better interpret our results, but also to report and discuss the limitations of the dHCP data and segmentation. 

\section{Material and methods}
\subsection{Neonatal MRI data and ground truth segmentation}
In this work, we evaluate the performance of the SynthSeg approach \cite{billot_synthseg_2023} on neonatal data using the third release of the publicly available developing Human Connectome Project (dHCP) dataset (http://www.developingconnectome.org/) \cite{edwards_developing_2022}. The dHCP dataset contains high-quality anatomical MRI scans of 885 neonates (age range from 26 to 45 weeks post-conception) acquired with both T1-weighted (T1w) and  T2-weighted (T2w) sequences with a 0.5 mm isotropic resolution on a 3T Philips scanner (see  \cite{edwards_developing_2022} for further information about acquisitions). 

\subsubsection{Ground truth based on volumetric segmentation: GT\_drawEM}
We first used the segmentation in 9 tissues provided by the dHCP consortium  \cite{makropoulos_developing_2018} ( \textbf{CSF} (Cerebrospinal fluid), \textbf{GM} (Cortical gray matter), \textbf{WM} (White matter), Background, Ventricles, \textbf{Cereb} (Cerebellum), \textbf{deepGM} (deep Gray Matter), \textbf{Bstem} (Brainstem), \textbf{HipAmy} (Hippocampi+Amygdala)). The segmentation is based on the multi-atlas method drawEM \cite{makropoulos_automatic_2014} applied to the T2w data. As mentioned by the authors, drawEM is very robust and efficient in most cases but may fail to capture the highly complex shape of the cortical geometry.  Extensive quality control was performed prior to the first release of the data, but localized inaccuracies remain. For instance, the authors reported that entire folds may be excluded from the automatic segmentation in 2\% of cases. As a consequence, it is important to remind throughout this study (and other works focusing on segmentation using this dataset) that the segmentations provided should be  considered as \textit{pseudo} ground truth, although the term “ground truth” is used for simplicity.  In this work, we merged the CSF and Ventricle labels into a single class (only for the evaluation) in order to avoid potential perturbations in the performance related to the tedious delineation between these two labels with similar intensity distributions. We refer to these ground truth segmentation maps as \textbf{GT\_drawEM}. 
\subsubsection{Ground truth based on surface reconstruction: GT\_surf}
\label{methodGT}
\begin{figure*}
	\centering 
	\includegraphics[width=\textwidth ]{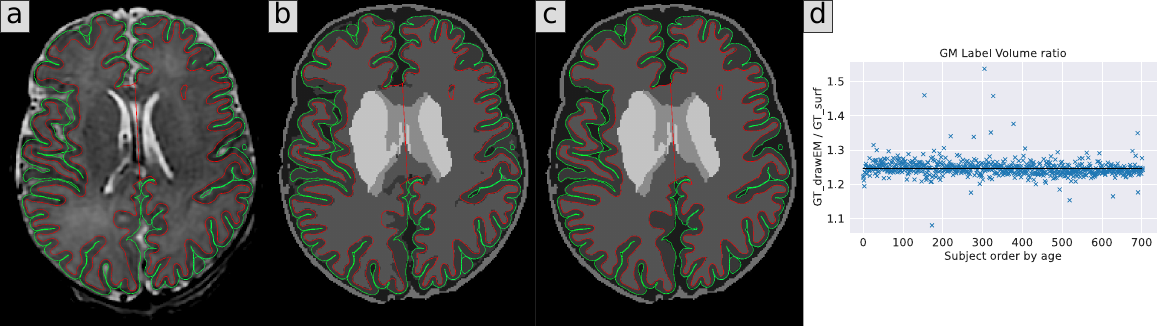}	
	\caption{Illustration of the two different ground truths used in our study. We overlay in red and green the internal and external GM surfaces resp. a) original T2w; b) label map from GT\_drawEM; c) label map from GT\_surf. d) Volume of GM from GT\_drawEM divided by the volume of GM from GT\_surf for each subject, ordered by age. The black line represents the mean value of 1.25, illustrating a 25\% increase of GM volume in GT\_drawEM compared to GT\_surf.}            
	\label{fig1}
\end{figure*}

We derived a second type of ground truth from the same images, based on the internal (white) and external (pial) cortical boundaries, represented as surfaces. Both surfaces are provided by the dHCP and are computed using the surface deformation tool introduced in \cite{schuh_deformable_2017}. Briefly, white-matter (internal) surface extraction is performed by fitting a closed, genus-0, triangulated surface mesh onto the segmentation boundary under constraints incorporating intensity information from the T2w, as well as controlling for surface topology and smoothness. The external (pial) surface is then obtained by deforming the internal surface outwards in order to fit the tissue boundaries \cite{makropoulos_developing_2018, schuh_deformable_2017}. From these internal and external surfaces surrounding the cortical tissue, we compute partial volume maps of the GM on the same 3D voxel grid as the T2w volume using a surface-based approach \cite{kirk_toblerone_2020}. We then obtain a binary segmentation of the GM by applying a threshold of 0.5. Finally, this different segmentation map for the GM is incorporated into the DrawEM label maps by replacing the original GM label and propagating the adjacent labels to preserve their topology. Compared to the original segmentation maps GT\_drawEM, all the structures remain identical except WM, GM, and CSF. We denote this second segmentation map as \textbf{GT\_surf}.\\
We illustrate these two types of ground truth segmentation maps in Fig. \ref{fig1}, with a plot showing the GM volume ratio (GT\_drawEM/GT\_surf) by subject, ordered by age. While the differences might look subtle visually on a single slice, we measured an average 25\% increase of GM volume in the GT\_drawEM compared to GT\_surf. This ratio is not influenced by the age of the baby. In this study, we use these two different, but both anatomically plausible, pseudo-ground truths to assess the influence of the definition of the segmentation map on the predictions of the models. 

\subsubsection{Head label}
The background label from drawEM maps contains only a thin layer surrounding the CSF, as the segmentations are computed on a brain-masked volume \cite{makropoulos_developing_2018}. Using only these labels for the generative process would limit the application to segment only skull-stripped data. We add other head tissue labels using the MIDA template \cite{iacono_mida_2015}, which contains 153 labels segmented from an adult MRI. We extract the extra-brain labels, which we grouped into 9 classes (dura mater, air, eyes, mucosa, muscle, nerves, skin, skull, and vessel) plus the background. These labels are combined with the 9 labels of the drawem9\_dseg volume after registering the MIDA template to each subject using the FIRST method from FSL \cite{jenkinson_global_2001} for the affine part, and reg\_f3d from NiftyReg \cite{modat_fast_2010} for the nonlinear part. We perform the label fusion to keep the original labels within the brain unchanged, and missing voxels outside the brain are set to air tissue. All these labels were used for the synthetic data generation but were then grouped into a single class (head) for the target objective.

\subsection{Generative model}
The key idea of the SynthSeg approach proposed in \cite{billot_synthseg_2023} consists in generating the entire training set as synthetic images from 3D segmentation labels, meaning that no real image is used for training. This is based on the assumption that the MR signal is homogeneous within each label. We implemented different transforms for simulating variations in tissue intensity, shape variability, and MRI artifacts (bias and noise) using a generative model detailed below. We further enriched the generative model from \cite{billot_synthseg_2023} by adding simulated motion artifacts and white matter inhomogeneity. The generative model was implemented using PyTorch and TorchIO \cite{perez-garcia_torchio_2021}. \\
\noindent \textbf{Random contrast}. The first step consists in generating an MRI volume from the label set by sampling the intensity of the voxels of each tissue from different Gaussian distributions, resulting in a synthetic MRI with a random contrast. The mean and the standard deviation of the Gaussian distribution of each tissue are sampled independently from uniform distributions U[0,1] and U[0.02, 0.1], respectively as in \cite{billot_synthseg_2023} (U[a, b] is the uniform distribution in the interval [a, b]). We implemented this process within TorchIO with the RandomLabelsToImage transform.\\
\textbf{Shape variability}. To generate synthetic MRIs with variations in brain anatomy from a limited number of subjects, we apply affine and non-linear deformations to the label set with nearest-neighbors interpolation. We use a composition of the following transforms from TorchIO: RandomAffine (scaling factor ~ U[0.9, 1.1] rotation ~ U[-20°, 20°], translation ~ U[-10mm, 10mm]), and RandomElasticDeformation (12 control points and a max displacement of 8mm). \\
\textbf{MRI artifacts}. We further augment the synthetic dataset by adding an intensity bias field and a global Gaussian noise. We use RandomBiasField, which simulates spatial intensity inhomogeneity with a polynomial function of order 3 and a maximum magnitude of 0.5, and RandomNoise which add a Gaussian random noise with 0 mean and a standard deviation sampled from U[5e-3, 0.1].

\begin{figure*}
	\centering 
	\includegraphics[width=\textwidth ]{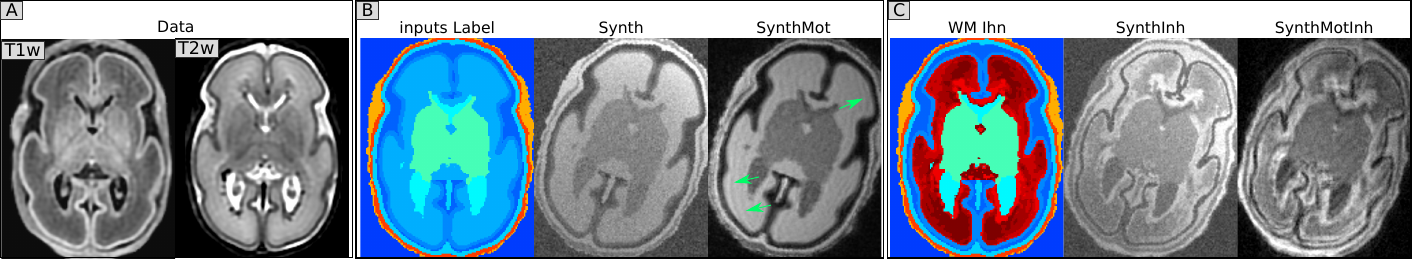}	
	\caption{Illustration of the synthetic datasets obtained from one individual data. A) the T1w and T2w MRI data from this subject; B) and C) show respectively the 2 different label maps (in color) used as input of the generative model (without and with additional labels in the WM to simulate inhomogeneities). For each label map, we show an illustrative example of augmented synthetic images, corresponding to the 4 synthesis-based models. Green arrows indicate subtle artifacts induced by motion simulation.}            
	\label{fig2}
\end{figure*}

\textbf{Motion simulation (Mot)}. We then extend the data augmentation beyond \cite{billot_synthseg_2023} by adding a RandomMotion transform to simulate subject motion during the MRI acquisition. We use our own implementation of the motion simulation introduced in \cite{reguig_global_2022}, which allows us to simulate a realistic time course of rigid head motion. As shown in Fig. \ref{fig2}, the motion simulation induces inhomogeneities in the different tissues because of the tissue mixing in the k-space induced by motion. We use a maximum displacement sampled from U[3, 8] (in mm) for the translation and U[3, 8] (degrees) for the rotation. Note that we force the background signal intensity to zero when generating motion artifacts to avoid mixing background intensity with the motion process. \\
\textbf{White Matter inhomogeneities (Inh)}. Inhomogeneities in the WM tissue are expected during the developmental period covered, and are visible in the MRI data. We adapted the method proposed by Billot \cite{billot_synthseg_2023} to account for such variations within a label: we subdivide WM into smaller sub-regions by clustering the T2w intensities, within the WM mask, using the Expectation Maximization algorithm \cite{dempster_maximum_1977}. We choose N regions ( $N \subseteq[2,3,4,5,6]$ ) in order to represent the inhomogeneities with different levels of granularity. Each subregion is then considered as a distinct tissue in the generative model (thus with a different random intensity) but they are regrouped for the segmentation objective in order to predict the whole WM.  (see Fig. \ref{fig2} c). Note that the term ‘transform’ is used for simplicity but is not adapted here since it is only a fixed modification of the input labels. Note also that real T2w data is used only for the generation of the label maps, but no real data is used for generating the synthetic images. 

Finally, an intensity normalization is performed to scale the min and max signal intensity between 0 and 1 for each synthetically generated dataset. This generative model is used to produce synthetic training sets based on the same ground truth segmentation maps for the following 4 synthesis-based models:\\
 \noindent \textbf{Synth}: SynthSeg method (same as Billot) with the following augmentation: random contrast, shape variability (affine and non-linear) and MRI artifacts (intensity Bias and noise)\\
 \textbf{SynthMot}: Synth enriched with motion simulation with a probability of 0.5\\
 \textbf{SynthInh}: Synth with extra labels within the white matter to simulate inhomogeneities (with a probability of 0.5)\\
 \textbf{SynthMotInh}: Synth with a combination of both WM inhomogeneities and simulated motion augmentations with a probability of 0.5 each \\
 \textbf{DataT2} (baseline): The performance of the 4 models based on synthetic training sets are compared with a baseline model defined as a UNet trained on real dHCP T2w acquisitions from the same 15 subjects. We apply the same data augmentation as for the synthesis-based models except for random contrast and Motion. We also add a random gamma augmentation to simulate slight variations in the intensity distribution \\

\subsection{Training and backbone architecture of the models}
\noindent \textbf{Training and testing sets}. The final sample of data used in this work was composed by selecting the images from the 709 scanning sessions of the dHCP with both T1w and T2w available among the 885 scanning sessions. 5 were excluded due to failure to generate the GT\_surf ground truth segmentation maps. Among the 176 sessions for which only the T2w was available (without the T1w acquisition), we selected 15 sessions uniformly distributed across the entire age range to generate the synthetic training sets for the models. In total, the data from 719 subjects from the dHCP data were used in this study: 15 for the training set and 704 subjects for the test set. We used the unprocessed T1w and T2w (no brain mask or bias field correction). \\
\textbf{Network architecture}. The network architecture used for all methods was the well-established 3D Unet architecture \cite{ronneberger_u-net_2015} with residual skip connections. We used five levels, each separated with either a max-pooling for the encoder path or an upsampling operation for the decoder part. All levels contained 3 convolution layers, with $3*3*3$ kernels. Every convolutional layer was followed by a batch normalization, a ReLu activation function, and a 10\% dropout layer, except for the last one, which was only followed by a softmax. The first block contained 24 feature maps and this number was doubled after each max-pooling and halved after each upsampling. This led to a total of 21,6 million parameters. \\
\textbf{Training}. All the models were trained with patches of size $128^3$. For each generated volume, we randomly selected 8 patches, sampled from a uniform distribution with the same probability of containing each structure. All models were trained with the average dice loss. Note that thanks to the fully convolutional nature of the Unet architecture, the inference was performed on the entire volume at native resolution. We used a batch size of 4 and the Adam optimizer with a learning rate of 1e-4. The training was stopped after 240 000 iterations. The training of each model took 6 days on an NVIDIA tesla V100 GPU ( http://www.idris.fr/).

\subsection{Quantitative measures and qualitative evaluation}
We report the binary dice, which is commonly used for segmentation evaluation, defined as $dice = 1 - 2 * (X * Y) / (X^2 + Y^2)$, where $X$ is the binary prediction of a given tissue and $Y$ is the ground truth label (already binarized). We also computed the average surface distance from MONAI \cite{cardoso2022monai}, but do not report this measure since it is fully consistent with the Dice score. We report the distribution of the Dice score across individuals separately for the different labels, as well as the distribution of the average of the Dice across all labels. In order to assess the potential effect of the age of the babies on the predictions, we also report the distribution of the Dice averaged over all structures computed in four age groups: 29 subjects in [26, 32[; 96 in [32, 36[; 183 in [36, 40[; 394 in [40 45]. To assess the robustness of the prediction to changes in image contrast, we computed for each tissue type the Pearson correlation between the volumes obtained from the predicted segmentation from either the T1w or T2w images, across the 704 subjects of the test set. An ideal, fully contrast-independent segmentation technique would produce almost identical segmentations from either T1w and T2w images and thus get a correlation value close to 1.\\
Visual assessment is critical to complement quantitative measures and better interpret the results of segmentation tools but is time-consuming and expertise-demanding. As a tradeoff, we focused our visual assessment on the GM, which is the most challenging anatomical structure to segment, and thus appropriate for assessing the variations in performances across the methods. We describe and illustrate our observations in combination with the quantitative measures for each of our experiments in the next section.

\section{Experiments and Results}
We designed three different experiments in order to address the following questions: 1) What are the performances of SynthSeg and our enriched versions compared to training on real data? 2) Is the high robustness with respect to variations in image contrast reported in \cite{billot_synthseg_2023} confirmed on neonatal MRI data? 3) How does the definition of the ground truth segmentation maps affect the performances? For each experiment, we provide both quantitative and qualitative assessments allowing us to interpret potential variations in the performances across the models. This extensive visual assessment enabled us to identify different types of limitations in the segmentation provided by the dHCP. We report in section \ref{resQC} our observations that we believe are important for future studies on this widely used dataset.
\subsection{Experiment \#1: evaluation of synthesis-based approaches on dHCP T2w dataset}
The aim of this experiment was to assess the performances of synthesis-based methods on neonatal brain MRI data and compare them to a learning strategy on real data in the absence of domain shift and with high-quality data. To this end, we used the following experimental setup:

\begin{itemize}[leftmargin=0.3 cm]
\setlength\itemsep{0em}
\item Training set: 15 subjects, ground truth = GT\_DrawEM, data used for DataT2: T2w
\item Testing set: 704 subjects, ground truth = GT\_DrawEM, prediction for all methods on T2w
\end{itemize}


\begin{figure*}[t!]
  \centering
  \includegraphics[width=\textwidth ]{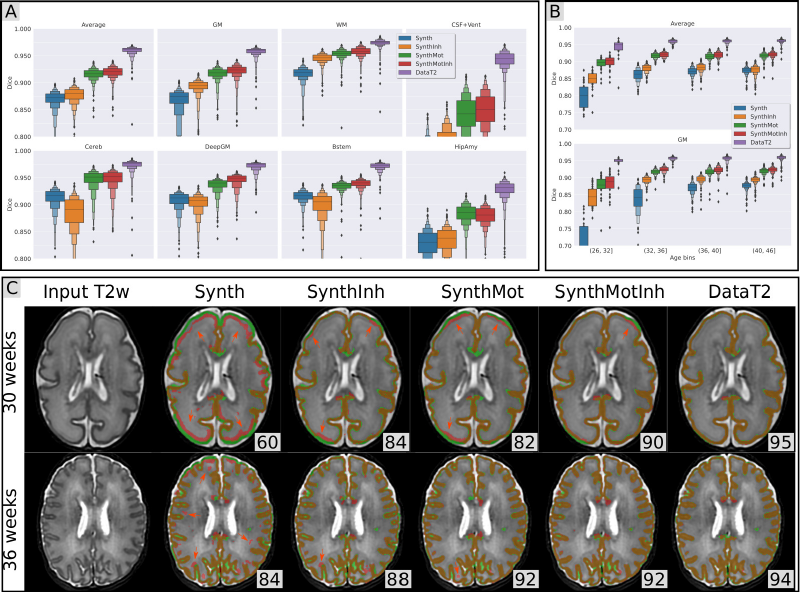}	
  \caption{A) Distribution across the 704 subjects of the test set of the Dice score for each structure, for the 5 models. B) Distribution of the Dice averaged across all structures, computed in four age groups: 29 subjects in [26, 32[; 96 in [32, 36[; 183 in [36, 40[; 394 in [40 45]. C) illustrations of the predicted GM for 2 subjects with different cortical folding magnitudes related to their age: sub-CC00657XX14 (30 weeks) and sub-CC00570XX10 (36 weeks). The numbers correspond to the dice score computed for the slice shown. Ground truth label (GT\_drawEM) is shown in green, and the predictions in red. Red arrows indicate local errors in the predictions}
            \label{fig3}
\end{figure*}

As can be seen in all of the plots of Fig. \ref{fig3}, the performance of the methods are ranked in a consistent order across all structures and across age groups. DataT2 is performing best with an average dice of 96 (and above 95 for all structures except CSF and Hip-Amy with respectively 94 and 92). The performance of the Synth model proposed by \cite{billot_synthseg_2023} is lower than the DataT2 model by 9 dice points on average (87). The differences are smaller (between 5 and 6 points) for WM/bstem/cereb/deepGM, but larger (~10 points) for GM and Hip-Amy and even 20 points for CSF+ventricle. Adding the motion augmentation greatly improves the performance, for all structures and age ranges. The difference between the SynthMot model and the DataT2 model is reduced by a factor of 2 for all structures, with an average dice of 92 (only 4 points difference with DataT2). Adding white matter inhomogeneity in the SynthInh model is also beneficial compared to the Synth, but the gain is mitigated: we observed an improvement for GM but not for Cereb DeepGM and Bsteam.  On the other hand the SynthMotInh model with both, motion and WM inhomogeneity, performs best (after dataT2) with a slight improvement compared to SynthMot.
Regarding the effect of age, we observe a drop in performance for all methods for the younger group (below 32 weeks). While the performance loss for DataT2 is limited, the Synth model shows the largest decrease in performance related to age, with a drop of 9 points. Adding motion augmentation and white matter inhomogeneities in SynthMotInh clearly mitigated this drop in the performance of the synthesis-based approach. 
The visual assessment showed that results are very consistent among subjects, with noticeable differences across methods on the first age bin. As illustrated in Panel C of Fig. \ref{fig3}, the predicted GM from the Synth model shows large errors with shifts of the GM prediction within the WM for the younger subjects. Those errors are largely fixed with the enriched synthesis-based models (SynthInh / SynthMot / SynthMotInh). The loss in performance of the synthesis-based models (even SynthMotInh) compared to DataT2 observed on Panel A and B corresponds to much more localized but clear errors.
For older subjects, the types of errors are different. The errors of the Synth model are restricted only to regions where marked signal inhomogeneities are present in the WM. Here also, those local errors are largely fixed with other synthetic alternatives. The lower dice scores compared to DataT2 are mainly due to subtle errors along the boundary between GM and WM or CSF. Careful visual inspection showed that the SynthMotInh prediction better follows image contrast than the GT\_drawEM. Regarding the model DataT2 (evaluated on T2w in this experiment) is performing almost perfectly at all ages: the predictions strictly follow the ground truth. The slight differences with GT\_drawEM are due to local errors of the ground truth, even for the first age bin. 
Overall, our observations are:
\begin{itemize}[leftmargin=0.3 cm]
\setlength\itemsep{0em}
\item DataT2 is highly accurate at every age even when trained on only 15 subjects.
\item Synth does not perform well, especially on younger subjects, with large regions of GM shifted within the WM.
\item The enriched synthesis-based models enable to fix most of the errors but local errors still occur especially for the younger subjects.
\item For older subjects, the predictions from SynthMotInh are visually accurate and better follow the underlying image contrast than the GT\_drawEM.
\item DataT2 does not make any obvious error (except for one subject), it reproduces the same tissue boundary as the ground truth and seems more robust to image noise than GT\_drawEM. 
We further examine the anatomical relevance of the predictions relative to GT\_drawEM in section \ref{resQC} below.
\end{itemize}

\subsection{Experiment \#2: robustness to variations in image contrast}
In the second experiment, we used the same trained model and changed the evaluation. Our aim was to assess the robustness of the models with respect to variations in the contrast of the image of the test set relative to the training set. We took the T1w image from the same individuals as an extreme change in image contrast relative to the T2w. We used the following experimental setup:
\begin{itemize}[leftmargin=0.3 cm]
\setlength\itemsep{0em}
\item Training set: same as Exp.\#1 (15 subjects, ground truth = \textbf{GT\_DrawEM}, data used for DataT2: T2w)
\item Testing set: 704 subjects, ground truth = GT\_DrawEM, prediction for all methods on \textbf{T1w}

\end{itemize}

\begin{figure*}[t!]
	\centering 
	\includegraphics[width=\textwidth ]{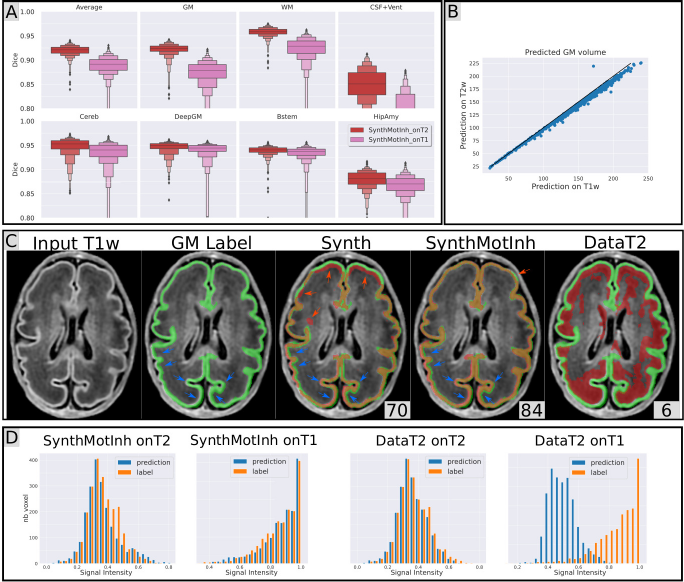}	
	\caption{ A) distribution across all subjects of the Dice score computed between GT\_drawEM and the predictions of the SynthMotInh model from either T2w or T1w, for the different structures. B) scatter plot of the GM volume computed from prediction of the SynthMotInh model. On the y-axis predictions are made from the T2w volumes and on the x-axis from the T1w volumes. C) illustration of the visual observations across the different methods. Ground truth label (GT\_drawEM) is shown in green, and the prediction in red. Blue arrows indicate regions with visible misalignment of the GT with regard to the T1w image. Red arrows indicate local errors in the predictions. D) histograms of the intensities of the T1w and T2w images within the predicted GM label (in blue) and GM GT (in orange).}            
	\label{fig4}
\end{figure*}

We show on Fig. \ref{fig4} detailed results for the best synthesis-based model from Experiment \#1 SynthMotInh and report the results for all the methods in a supplementary csv file \footnote{\url{https://github.com/romainVala/Synthetic_learning_on_dHCP}}. All panels show that DataT2, as expected, failed to predict on T1w inputs, with an average Dice below 10. The histograms within the GM shown on Panel E confirm that the dataT2 model learned the correspondence between the labels and the intensity distribution: the predicted GM from the T1w image (in blue) corresponds to voxels that have the same intensity range as the intensity in the GM from T2w image. On the contrary, the histograms from Panel D show that the SynthMotInh model gives very consistent segmentations from both T2w and T1w inputs. Indeed, the intensity distributions within the predicted GM from T1w and T2w images are very different and match the ground truth distribution well in both cases. Panel B shows the strong linear correlation between the volumes of the GM obtained from the SynthMotInh predictions from the T1w and T2w images across the 704 subjects of the test set. The Pearson correlation is above 0.99. This plot also shows a slight deviation of the data compared to y=x, indicative of a slightly larger estimated GM volume on T1w compared to T2w (5\% on average).\\
The visual assessment was critical for this experiment, as illustrated in Panel C. First, we observe that the Synth model suffers from the same limitations as in Exp \#1. Regarding the SynthMotInh model, we observe that the predictions on T1w are visually as good as the one from T2w, which is consistent with the high correlation shown on Panel B, but inconsistent with the drop of 4 points of Dice on average across all structures shown on Panel A. More specifically, robustness is excellent for DeepGM, Bstem, Cereb, and Hip-Amy with a decrease in performance of only 0.7, 1.7, 2.4, and 2 points of Dice respectively. The Dice is however lower for GM, WM, and CSF+Ventricle (loss of 5.4, 4, and 10 resp.). The same trends are observed for the other synthesis-based models. \\
Careful visual assessment enabled us to observe that the disagreement between prediction and GT is mostly due to residual misregistration between T1w and T2w images from the dHCP dataset. This is visible on Panel C for SynthMotInh, with a slight shift in the location of GM especially in the left posterior region of the brain. The impact of such residual misregistration on the Dice scores is stronger for external tissues (GM, WM, and CSF+Ventricles) than for deep structures, as observed in Panel A. We report further observations regarding the impact of misregistration on our evaluation in subsection \ref{resQC}. Overall, our observations are:
\begin{itemize}[leftmargin=0.3 cm]
\setlength\itemsep{0em}
\item The DataT2 model can not generalize to other contrasts;
\item The synthesis-based models perform equally well on both modalities; 
\item Remaining differences in Dice are mostly due to misregistration between T1w and T2w and not to segmentation errors. 
\end{itemize}

\subsection{Experiment \#3: Influence of the definition of the ground truth}
The aim of this experiment was to assess the influence of the definition of the ground truth on the performance of the models. We computed another ground truth for the GM derived from the cortical surfaces GT\_surf, as explained in section \ref{methodGT} (all other tissues remain the same, except WM and CSF). We rerun the same experiments with this new ground truth for two models only: SyntMotInh and DataT2. We used the following experimental setup:
\begin{itemize}[leftmargin=0.3 cm]
\setlength\itemsep{0em}
\item Training \& testing \#1: ground truth = GT\_DrawEM,  prediction on T2w compared to GT\_DrawEM (same model as Exp\#1)
\item Training \& testing \#2: ground truth = GT\_Surf, prediction on T2w compared to GT\_Surf
\item The influence of the change in the ground truth is assessed by comparing the predictions from the two training sessions. 
\end{itemize}

\begin{figure}
	\centering 
	\includegraphics[width=0.5\textwidth ]{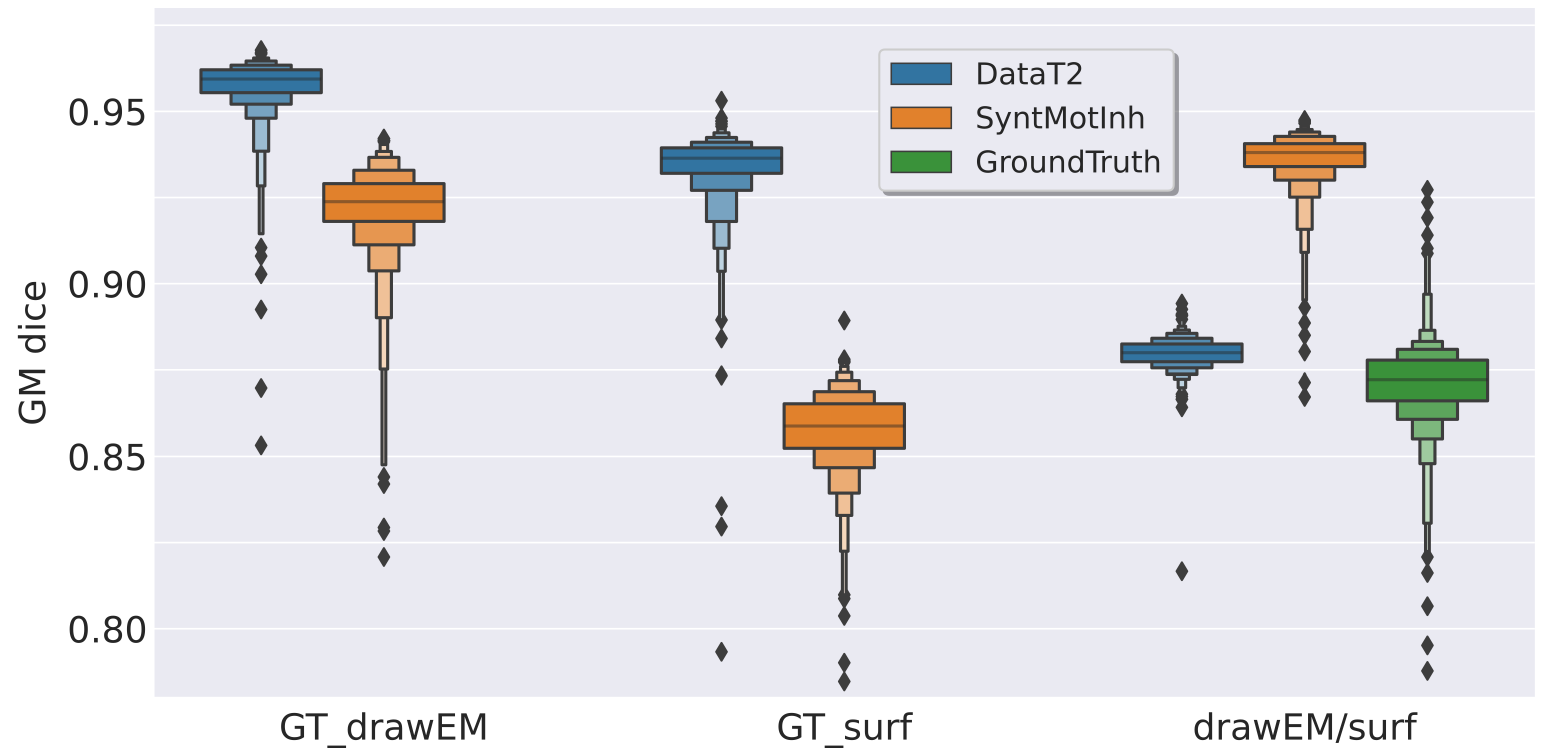}	
	\caption{ Dice computed between predictions and ground truth. The columns GT\_DrawEM show the dice obtained when the models are trained and evaluated using the GT\_DrawEM ground truth. (same values as Fig. \ref{fig2}.) The columns GT\_surf show the measures when the models are trained and evaluated with the GM derived from the surfaces. For the columns drawEM/surf, the Dice is computed between the predictions of the model trained with GT\_drawEM and the predictions of models trained with surf GT\_surf. For comparison, we also show the Dice between the two ground truths (in green).}            
	\label{fig5}
\end{figure}

The results from this experiment for the GM are reported on Fig. \ref{fig5}. The first observation is the clear impact of the ground truth on the performance of the models. We observe a drop of 2 points for DataT2 when using GT\_surf instead of GT\_DrawEM (from 95 to 93). The drop is even larger for the SynthMotInh model: 6 points (from 92 to 86). On the other hand, when we look at the consistency between the predictions from the two training sessions (column drawEM/surf), we observe a dice of 88 for the DataT2 model, which is very close to the dice between the two ground truths. In contrast, the Dice value of 94 for the SynthMotInh model indicates that the predictions are much more consistent, showing a lower influence of the type of ground truth used in the training set.\\
Visual assessment enabled us to better interpret the drop of 2 points of Dice for Data T2 model trained (and evaluated) with GT\_surf compared to the same model trained on GT\_drawEM. Indeed, this drop in performance is not due to a more difficult task or less accurate predictions, but to local inaccuracies in GT\_surf. We observed focal errors in GT\_surf that are related to bad positioning of the white or pial surfaces, probably due to the balance between topology correction and data attachment terms in the surface deformation algorithm \cite{schuh_deformable_2017} These observations also explain the spread in the distribution of Dice computed between the two ground truths (in green on Fig. \ref{fig5}). Note also that the spread is reduced for the prediction of DataT2 models (third column in blue compared to green), demonstrating a better robustness of DataT2 predictions compared to the both ground truths.

In summary, the DataT2 model learns very well the ground truth whatever its definition, even with only 15 subjects. The synthesis-based model is less impacted by a change in the definition of the ground truth labels used for the generative model. 

\subsection{Detailed assessment of image quality and anatomical relevance of the segmentation provided by the dHCP}
\label{resQC}

\begin{figure*}[t!]
	\centering 
	\includegraphics[width=\textwidth ]{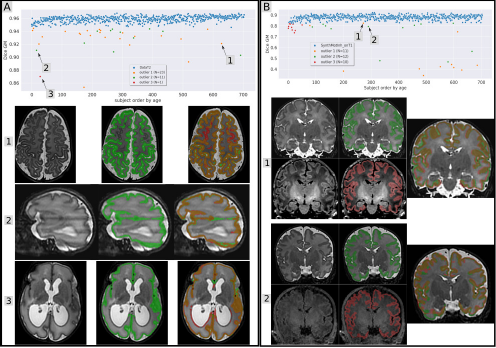}	
	\caption{ A) Dice score of DataT2 model (evaluated on T2w) for all subjects ordered by the age of the subjects. We visually checked the 35 outliers, grouped them into three categories, and provided an example for each type. (The examples shown below are indicated by a black arrow). B) Dice score of the SynthMotInh GM predictions (evaluated on T1w) for all subjects, after excluding the 35 outliers from panel A. We visually checked the 33 outliers and grouped them into three categories. Chosen examples are shown by a black arrow. Ground truth labels (GT\_drawEM) are shown in green, and the predictions in red.}            
	\label{fig6}
\end{figure*}

The anatomical validity of ground truth is rarely discussed in the deep learning literature, but it plays an important role. In this work, we used the segmentation provided by the dHCP consortium as one of our ground truths (GT\_drawEM). These segmentation maps were obtained using automated image processing as described in \cite{makropoulos_developing_2018}. The segmentation pipeline was optimized for instance by modeling additional tissue classes to account for inhomogeneity in WM. Such segmentations are of great value and our study would not be feasible without such material, as well as many other publications. In this section, we report additional observations from our extensive visual assessment that might serve for future works based on this dataset. We visualized systematically all the images corresponding to potential outliers, i.e. for which the Dice score was far from the mean. For comparison, we also visualized randomly picked images to assess the average performance.\\
As illustrated in Panel A of Fig. \ref{fig6}, the anatomical relevance of the predictions from DataT2 is better than GT\_drawEM. We identify three types of outliers: Outlier type 1 (N=23) corresponds to obvious, relatively large errors of the GT\_drawEM, despite the high quality of the T2w image; Outlier type 2 (N=11) corresponds to lower quality T2w data for which GT\_drawEM was affected by artifacts, while the prediction from the DataT2 model looks much more anatomically relevant; Outlier type 3 (N=1) corresponds to the only subject for which the prediction from DataT2 shows obvious errors with False positive GM prediction near the ventricle. Note that this subject was classified as pathological by the dHCP consortium (radiological\_score=5), which suggests that the visually enlarged lateral ventricles likely correspond to very large variations with respect to the normal brain configuration. \\
In Panel B of Fig. \ref{fig6}, we report our observations relative to the quality of the T1w versus T2w data from the dHCP. We visually checked the 33 outliers from the distribution of Dice score obtained for the SynthMotInh predictions of GM from T1w images and we identified three types: Outlier type 1 (N=11) corresponds to high-quality images but with obvious misregistration between T1w and T2w. Note that we show in panel B 1) the outlier with the highest dice score (above 0.8) but we observed 8 subjects with a dice lower than 0.5 due to obvious misregistration; Outlier type 2 (N=12) corresponds to subjects with bad quality T1w images, for which a low overlap with the GT is expected; Outlier type 3 (N=10) corresponds to very young subjects, for which the SynthMotInh model was less accurate (on both contrasts) than for the older ones. We do not illustrate this configuration since it is already shown on Fig. \ref{fig3} and \ref{fig4}. The misalignment and quality issues from outlier types 1 and 2 explain the loss of Dice in Exp \#2 since the GT\_drawEM has been defined from the T2w volumes only.

Overall, the visual assessment of the outlier showed cases with large errors in the GT. Either because of failure of the drawEM pipeline, or because of mis-coregistration issues. In addition, for both experimental examples we also found outliers due to the poor image quality.

\section{Discussion}
\subsection{Synthetic learning models require adaptations to perform on neonatal MRI}
The synthetic approach proposed by Billot et al. (model Synth) performed worse than we anticipated. However, this low performance only affects the GM of the youngest subjects (i.e. those with unfolded GM). The method is more effective for older subjects, which agrees with its good performance for the adult brain \cite{billot_synthseg_2023, billot_robust_2022}. However, this finding is unexpected since the GM segmentation might appear simpler to perform on brains without convolution. Furthermore those errors occur in regions with high image quality, with a clear contrast between GM and surrounding tissues. We hypothesize that this failure mode is due to a too simple generative model that cannot account for the variations in intensity within the immature WM, which are much larger in this dataset compared to adult brains. Our results show that adding motion augmentation in the generative process significantly improves the performance, as shown by the SynthMot results. The motion simulation mixes different tissues and can lead to localized artifacts (illustrated on Fig. \ref{fig2}) that are qualitatively similar to the inhomogeneities in the white matter induced by maturation, with a spatial pattern in layers propagating inward from the GM \cite{pogledic_subplate_2020}. Therefore, the gain in performances might be interpreted as a positive side effect rather than an anatomically relevant data augmentation. Anyway, these observations confirm the statement from \cite{billot_synthseg_2023,tobin_domain_2017, tremblay_training_2018} that in the domain randomization approach, the key is to generate enough variations to cover the expected variations from real data, even if the generative process does not perfectly model the real data generation. \\
In addition to data augmentation with simulated motion, we also explored the alternative solution of explicitly modeling heterogeneity in WM by decomposing it into subregions (between 2 and 6). Our results suggest that this approach is less efficient than the proposed motion augmentation but still beneficial. Further work is needed to fully understand the potential of this strategy, as performance gains could be achieved by refining the number of subregions or by incorporating additional anatomical priors.

\subsection{Robustness to variations in image contrast}
Our results confirm that the DataT2 model did learn a correspondence between the spatial location and the underlying intensity in the image, as expected. Improving the generalization properties of the deep learning models is a very active topic, with various strategies investigated in parallel such as e.g. \cite{tomar_self-supervised_2022, zhao_data_2019, ouyang_causality-inspired_2022}. In this article we did not enter into evaluating these methods since in the context of early brain development, variations in image contrast have a biological meaning and might not be considered as a domain adaptation problem. Indeed, robustness to variations in image contrast is a key feature in this context, which is different from generalization between two predefined domains.\\
In our experiments, we consider the variations in contrast between T1w and T2w as a prototypal, extreme change. We confirm the contrast-agnostic properties of synthesis-based models with highly consistent predictions from either T1w or T2w images. We argue that the residual difference of 5 points of Dice for GM between predictions from T1w and T2w is not due to inaccurate predictions. We identified three main factors explaining this finding. First, residual misregistration between the T1w and T2w images from the same subjects are clearly present in the dHCP dataset. Second, different artifacts might affect the two acquisitions, impacting the predictions differently but inducing systematically a reduction of the Dice score. Third, the T1w and T2w image contrasts may be affected differently by brain maturation at the cellular level \cite{croteau2016examining}. Despite these uncontrolled sources of variance, the very high correlations across tissue volumes computed from T1w versus T2w confirm the robustness of the proposed SynthMotInh segmentation to variations in image contrast. \\
The residual mis-registration issues we report in Section \ref{resQC} are somewhat inconsistent with the dHCP image processing pipeline description \cite{makropoulos_developing_2018}. The authors noted that gradient non-linearity correction was not necessary, and reported that rigid co-registration was effective. However, this study was conducted on the first release of the dHCP dataset, which contains 465 subjects, whereas we included 704 subjects from the third release.  Through our careful visual assessment, we observed large and obvious residual mis-registration errors for at least 11 subjects. Therefore, we believe that this dataset is affected by mis-registration for a larger proportion of individuals. Our results suggest that comparing the segmentation predicted from a synthesis-based model from T1w and T2w is effective to detect mis-registration. This could be used to guide the registration between these two modalities. This is consistent with the recent study by Iglesias which introduced a robust multicontrast affine registration approach based on synthetic learning segmentations \cite{iglesias_ready--use_2023}.

\subsection{Robustness of synthesis-based approaches relative to the definition of the ground truth}
In general, the quantitative evaluation of a supervised segmentation method, such as dataT2 in this work, measures the network's ability to learn the ground truth from the training set (i.e. to learn the mapping between an image and the corresponding label map), regardless of the quality of the ground truth. The high DICE scores observed for dataT2 in Exp \#1 confirms this capacity to learn very accurately the ground truth. Consequently, the predictions are highly dependent on the ground truth, and any systematic bias affecting the ground truth would be learnt by such models. The bias would in turn affect the predictions. This was clearly confirmed by the result of Exp \#3 where the predictions of DataT2 models (GT\_drawEM and GT\_surf) show the same dice score (0.85) as the labels.\\
In contrast, the synthetic approach does not learn the relationship between image intensities and segmentation maps. By design, the synthetic framework relies on a generative model to simulate images, which induces an exact correspondence between the boundaries of the structures of interest and image intensities in the training set. This could explain the robustness of the predictions from the SynthMotInh method with respect to variations in the ground truth (see Exp \#3). Synthesis-based models are thus less biased by the quality of the ground truth than classical supervised models, which constitutes a new path to define boundaries between adjacent tissues, a major bottleneck for quantitative analysis.

\section{Conclusion}
In this work, we confirmed that synthetic learning for data segmentation offers key advantages compared to the classical strategy based on real data. In the context of newborn brain MRI, specific signal inhomogeneities affect the performance of the previously proposed synthetic approach. Our enriched generative model with motion simulation greatly improved the predictions, enabling contrast-agnostic segmentation of neonatal brain MRI. In addition, the synthesis-based models are not biased toward the specific intensity distributions of the training set, and the relationship between the geometry of the tissue and the intensity distribution is better controlled than with real data. Furthermore,  synthesis-based models can be trained using a very limited amount of manual segmentation examples. All these features will provide critical performance improvements in the perspective of large multi-site studies and clinical applications.

\section*{Acknowledgements}
\ssmall
This work was granted access to the HPC resources of IDRIS under the allocation 2022-AD011011735R3 made by GENCI.  The research leading to these results has also been supported by the ANR AI4CHILD Project, Grant ANR-19-CHIA-0015, the ANR SulcalGRIDS Project, Grant ANR-19-CE45-0014, the ERA-NET NEURON MULTI-FACT Project, Grant ANR-21-NEU2-0005 and the ANR HINT Project, Grant ANR-22-CE45-0034 funded by the French National Research Agency. dHCP Data were provided by the Human Connectome Project, WU-Minn Consortium (Principal Investigators: David Van Essen and Kamil Ugurbil; 1U54MH091657) funded by the 16 NIH Institutes and Centers that support the NIH Blueprint for Neuroscience Research; and by the McDonnell Center for Systems Neuroscience at Washington University.

\bibliographystyle{apalike}






\end{document}